# FML-BASED DYNAMIC ASSESSMENT AGENT FOR HUMAN-MACHINE COOPERATIVE SYSTEM ON GAME OF GO

**CHANG-SHING LEE\* MEI-HUI WANG, SHENG-CHI YANG**
*Department of Computer Science and Information Engineering, National University of Tainan, Tainan, Taiwan*
*\*leecs@mail.nutn.edu.tw, mh.alice.wang@gmail.com, skymini4910@gmail.com*

**PI-HSIA HUNG, SU-WEI LIN**
*Department of Education, National University of Tainan, Tainan, Taiwan*
*hungps@mail.nutn.edu.tw, swlin0214@mail.nutn.edu.tw*

**NAN SHUO, NAOYUKI KUBOTA**
*Dept. of System Design, Tokyo Metropolitan University, Japan*
*shuo-nan@ed.tmu.ac.jp, kubota@tmu.ac.jp*

**CHUN-HSUN CHOU, PING-CHIANG CHOU**
*Haifong Weiqi Academy, Taiwan*
*chouchunhsun@gmail.com, cabon1224@hotmail.com*

**CHIA-HSIU KAO**
*Department of Computer Science and Information Engineering, National University of Tainan, Tainan, Taiwan*
*tsubxxx@gmail.com*

In this paper, we demonstrate the application of Fuzzy Markup Language (FML) to construct an FML-based Dynamic Assessment Agent (FDAA), and we present an FML-based Human–Machine Cooperative System (FHMCS) for the game of Go. The proposed FDAA comprises an intelligent decision-making and learning mechanism, an intelligent game bot, a proximal development agent, and an intelligent agent. The intelligent game bot is based on the open-source code of Facebook's Darkforest, and it features a representational state transfer application programming interface mechanism. The proximal development agent contains a dynamic assessment mechanism, a GoSocket mechanism, and an FML engine with a fuzzy knowledge base and rule base. The intelligent agent contains a GoSocket engine and a summarization agent that is based on the estimated win rate, real-time simulation number, and matching degree of predicted moves. Additionally, the FML for player performance evaluation and linguistic descriptions for game results commentary are presented. We experimentally verify and validate the performance of the FDAA and variants of the FHMCS by testing five games in 2016 and 60 games of Google's Master Go, a new version of the AlphaGo program, in January 2017. The experimental results demonstrate that the proposed FDAA can work effectively for Go applications.

*Keywords*: Fuzzy markup language; prediction agent; decision support engine; robot engine; FAIR darkforest Go engine

## 1. Introduction

Go is an ancient game that has been widely played for more than 2500 years. Two players, Black and White, take turns placing their stone on an empty intersection, the objective of which is to surround the opponent's territory. There are 381 positions to play for on this 19×19 board game.[1] The player who controls the most territory at the end of the game wins the game. Additionally, a handicap is typically given to offset the strength difference between Black and White.[1] The level of amateur Go players is ranked by Kyu (K) and Dan (D), where 1K and 7D are the strongest level for Kyu and Dan levels, respectively. Professional Go players are also ranked by Dan (P), where 9P is the strongest level for professionals.[2]

The game of Go has been described as the "Mt. Everest" of artificial intelligence (AI) because of its high complexity and uncertainty, especially compared with other board games.[2] Games have served as an ideal benchmark for studying AI and computational intelligence.[3] Moreover, the course of game play in Go is sensitive to small changes; for example, adding or removing one stone could alter the life and death situation of a large group of stones, and thus change the game result.[4] Raonak Uz-Zaman[31] proposed two approaches to exploring reinforcement learning for the implementation of a Go computer program. The first approach combines the temporal difference method with the mixture-of-experts architectures for neural networks to learn reasonable Go evaluation functions for 9 × 9 boards. The second approach features a hybrid, modular architecture that incorporates search techniques with machine-learning algorithms. This hybrid system of neural networks and AI used in 19×19 Go is suitable for handling complex search and decision problems.

Monte Carlo tree search (MCTS) has had a profound effect on AI, particularly in computer games and Go.[2, 5] In 2009, Silver et al.[34] developed two algorithms for balancing the simulation policy according to a descending gradient in order to maintain an accurate spread of simulation outcomes. Programs such as MoGo[37]/MoGoTW, Crazy Stone, Fuego,[38] Many Faces of Go, and Zen were shown to be competitive at the professional level for 9×9 Go and at amateur Dan strength on a 19×19 board around 2010, and this was attributed to the use of MCTS .[39] Gelly and Silver[35] proposed two extensions to the MCTS algorithms, namely the rapid action value estimation algorithm and heuristic MCTS, and applied them to the Go program MoGo, achieving a significant performance improvement in 9×9 Go. Clark et al.[6] and Maddison et al.[7] have trained deep convolutional neural networks (DCNNs) to directly present and learn knowledge on Go and then predict the moves of expert Go players. Tian and Zhu[4] developed the computer Go program Darkforest, for which a DCNN was trained to predict the next top-$k$ moves. The combination of a DCNN with MCTS in Google's AlphaGo can approach the skill level of top professional Go players.[2, 8, 29, 30] Additionally, Google's Master Go program, a new version of AlphaGo, won 60 online games again top professional Go players in January 2017.[29, 30, 40] Moreover, AlphaGo beat top-ranked Go player Ke Jie in May 2017.[36, 40] In addition to Go, Xu et al.[9] presented a DCNN-based feature learning algorithm that was originally designed to automatically segment from digitalized tumor tissue microarrays. Zhang et al.[10] used

DCNN features in diverse visual recognition tasks. LeCun *et al.*[11] used gradient-based learning algorithms for document recognition. Yang *et al.*[12] proposed DropSample for large-scale online handwritten Chinese character recognition.

According to previous research,[32] there are four key considerations pertaining to advancements in AI. First, human learning is distinguished by the range and complexity of skills that can be learned and the degree of abstraction that can be achieved relative to other species. Second, discoveries in developmental psychology and machine learning are converging on new computational accounts of learning. Third, the objective of machine learning is to develop computer algorithms and robots that improve automatically through experience. Finally, psychology, neuroscience, machine learning, and education have contributed to a new science of learning. As data sets become larger and more complex, it is becoming increasingly difficult to analyze and extract inferences. Software developers must respond to new approaches to analyzing data in high-dimensional spaces by adopted pattern-searching algorithms used in statistics and machine learning.[33] In the present paper, we introduce the concept of the zone of proximal development (ZPD) to Go learning. The ZPD is the difference between what a learner can do without assistance and what he or she can do with assistance.[13] The ZPD concept, developed by Soviet psychologist Lev Vygotsky, is widely used in research on children's mental development in education. It posits that a child can follow an adult's example to develop the ability to do certain tasks without step-by-step help.[13] The role of education is to give the learner experiences that are within their ZPD, thereby encouraging and advancing their individual learning.[13]

The novelty and contributions of this paper are as follows: (1) Dynamic assessment (DA) is an assessment approach that blends instruction into assessment, and its basic framework includes pretesting, teaching, and posttesting.[14, 15] This paper presents a fuzzy markup language (FML)-based dynamic assessment agent (FDAA) with an intelligent decision-making and learning mechanism. Players are shown the simulation number, win rate, and next-moves prediction output by the intelligent game bot, which is based on the open-source code of Facebook' Darkforest Go, thus enabling adaptive predictions of game play. (2) The proposed agent combines the DA concept with item response theory[16] to offer Go players in-game suggestions and a postgame summarization. (3) A proximal development agent that combines the ZPD concept with related fuzzy sets for human–machine cooperative systems on the game Go. (4) FML is used to construct a knowledge base and rule base for an intelligent agent and proximal development agent. The FML-inferred linguistics could be "Black has an obvious advantage," "Black has a possible advantage," "White has an obvious advantage," "White has a possible advantage," or "both are in an uncertain situation," as determined by the MCTS simulation number, win rate prediction, and the matching degree of top-move rate prediction. (5) The intelligent agent retrieves the predicted next top-five moves from the proximal development agent and then provides feedback to players by giving a linguistic summarization.

The remainder of this paper is organized as follows: Section 2 outlines the preliminary concepts discussed in the paper, including a brief introduction on education assessment and FML, to help the reader understand and follow the concepts discussed herein. Section 3 describes the FDAA for the game of Go, the open-source code adopted from Facebook's

Darkforest Go, and the representational state transfer (RESTful) application programming interface (API) mechanism. The proximal development agent and intelligent agent are introduced in Section 4. In Section 5, the experimental results for applications to the game of Go are provided. Finally, conclusions are given in Section 6.

## 2. Preliminary Concepts

This section describes the preliminary concepts, including item response theory (IRT), assessment application, and FML.

### 2.1. *Item Response Theory and Assessment Applications*

IRT provides a one-, two-, or three-parameter logistic (1PL, 2PL, or 3PL, respectively) model[16] for adaptive assessment. It also features parameter invariance and information function. Each item has three parameters: difficulty, discrimination, and guessing when using the 3PL model. Only the item's difficulty is considered when the 1PL model is used. Student abilities are estimated according to their response patterns and the adopted model.[16] In an adaptive assessment, when an examinee answers correctly, a more difficult item is then be selected as his or her next item to respond to. By contrast, an easier item is presented when an incorrect answer is provided.[16, 17]

Assessment plays a crucial role in every learning and teaching activity.[18] There has been considerable research on adaptive assessments; for example, Lazarinis *et al.*[18] presented a personalized adaptive Web testing system based on learners' knowledge and objectives. Lee *et al.*[17] proposed a type-2 fuzzy set–based adaptive linguistic assessment system for the semantic analysis and human performance evaluation of the game Go. In the present paper, we adopt Go as a case study to assess the player performance as Black or White. The number of MCTS simulations could be regarded as an item's difficulty because it is adjusted to the strength of the opponent.[17] If the level of Black and White is similar, the numbers of MCTS simulations will be close to each other and the win rate will be maintained at approximately 0.5. Additionally, during play, players are given a hint (the predicted next top-five moves from Darkforest) as a reference through a computer or a robot. Therefore, this paper adopts the number of MCTS simulations, win rate, and top-move rate to assess player performance.

### 2.2. *Fuzzy Markup Language*

Many real-world applications with a high level of uncertainty have demonstrated the favorable performance of fuzzy sets. The IEEE 1855-2016 Standard for FML was approved by the IEEE Standards Associations (IEEE-SA) in 2016.[19, 20, 21] This standard provides designers of intelligent decision-making systems with a unified, high-level methodology for describing system behaviors on the basis of rules based on human domain knowledge.[21-24] The W3C XML Schema Definition is used to define the syntax and semantics of FML-based programs.[21-24] A considerable number of research applications have been based on FML and genetic FML such as an ambient intelligence environment,[24] a diet linguistic recommendation mechanism,[25] and an emotional expression mechanism for games of computer Go.[26] Table 1 provides a brief introduction to FML (for further details, refer to the IEEE Standard for FML[21]).

Table 1.    Brief introduction to FML.[21]

| |
|---|
| • Element *fuzzySystem*, containing elements *knowledgeBase* and *ruleBase*, is the top-level element of the schema of the FML specifications. |
| • Element *knowledgeBase* defines the knowledge base of the fuzzy system which contains *fuzzyVariable* or other elements like *tsukamotoVariable* or *tskVariable*. |
| • Element *fuzzyVariable* has nine attributes including *name*, *scale*, *domainLeft*, *domainRight*, *type*, *accumulation*, *defuzzifier*, *defaultValue*, and *networkAddress*. |
| • Element *fuzzyTerm* defines a linguistic term describing the fuzzy variable. |
| • Element *ruleBase* could be element *mamdaniRuleBase*, *tsukamotoRulBase*, *tskRuleBase*, or *anYaRuleBase*. For example, the element *mamdaniRuleBase* has five attributes to describe the constructed fuzzy rules of a mamdani-based fuzzy logic system. |
| • Element *rule* contains one element *antecedent* representing the antecedent part of a rule and one element *consequent* representing the [THEN-ELSE] part of a rule. |

## 3.  FML-based Dynamic Assessment Agent for Game of Go

Google AlphaGo successfully combines a DCNN, including policy and value networks, with MCTS.[8, 29, 30] In this section, we present an FML-based human–machine cooperative system (FHMCS) for Go player playing as well as learning and dynamic assessment with an intelligent game bot. The proposed structure is first introduced, and then the intelligent decision-making and learning mechanism is described. Finally, we present the intelligent game bot.

### 3.1.  *FML-based Dynamic Assessment Agent for Game of Go*

Figure 1 shows the structure of the proposed FDAA combined with the intelligent game bot. There are four mechanisms in the proposed platform, including (1) an intelligent game bot, with the open-source code for Facebook's Darkforest Go[4] and a RESTful API mechanism; (2) a proximal development mechanism with a dynamic assessment mechanism, a GoSocket mechanism, and an FML engine; (3) an intelligent agent with a GoSocket engine, a fuzzy knowledge base, and a summarization agent; and (4) an intelligent decision-making and learning mechanism with a decision-making mechanism and a real-time learning mechanism. The proposed FDAA is briefly described as follows:

- The game players play Darkforest through the decision-making mechanism and real-time learning mechanism, which give the real-time number of MCTS simulations and win rate of the predicted top-five next moves based on the Darkforest DCNN model. Meanwhile, the intelligent decision-making mechanism also stores the game play information in a database. The intelligent game bot is responsible for communicating with human Go players and Darkforest through the RESTful API mechanism by using Go Text Protocol (GTP) commands.
- The proximal development mechanism triggers the GoSocket mechanism to query the newest move to display changes in the number of MCTS simulations and win rate. Meanwhile, the FML engine sends the inferred real-time game situation analysis back to the GoSocket mechanism based on the WebSocket protocol.
- The intelligent agent provides the game players with a performance evaluation and game summarization in linguistic descriptions. The game players receive the recommended positions of the predicted top-five next moves through the intelligent game bot and the intelligent decision-making and learning mechanism. They also obtain an overall game summarization from the intelligent agent.

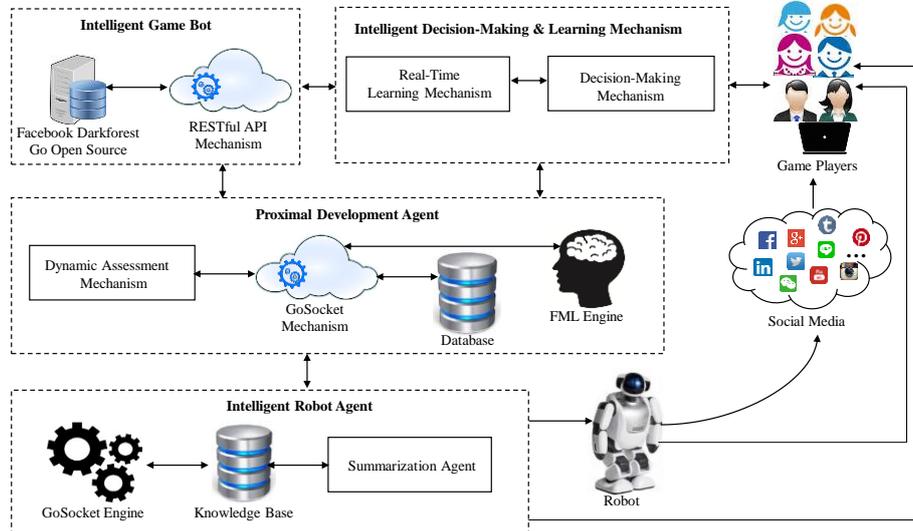

Fig. 1.     Structure of the proposed FML-based Dynamic Assessment Agent (FDAA) combined with the intelligent game bot.

### 3.2. *Intelligent Decision-Making & Learning Mechanism with Facebook Darkforest Go*

This subsection briefly describes the FHMCS for the game of Go and presents an example of a human vs. computer Go competition funded by the IEEE Computational Intelligence Society.[2] Owing to AlphaGo's impact on developments in computational intelligence,[2, 8] the human vs. computer Go competition at the 2016 IEEE World Congress on Computational Intelligence (IEEE WCCI 2016) included the concepts of the human–machine cooperation to popularize Go playing (http://oase.nutn.edu.tw/WCCI2016/). Chun-Hsun Chou (9P, Taiwan), Ping-Chiang Chou (6P, Taiwan), and Yi-Min Hsieh (6P, Japan) were invited to join the demonstration games of human–machine cooperation. Figure 2 shows the structure of the demonstration game among humans, Darkforest, and the intelligent game bot. Table 2 provides information on how to operate the demonstration game in the event.

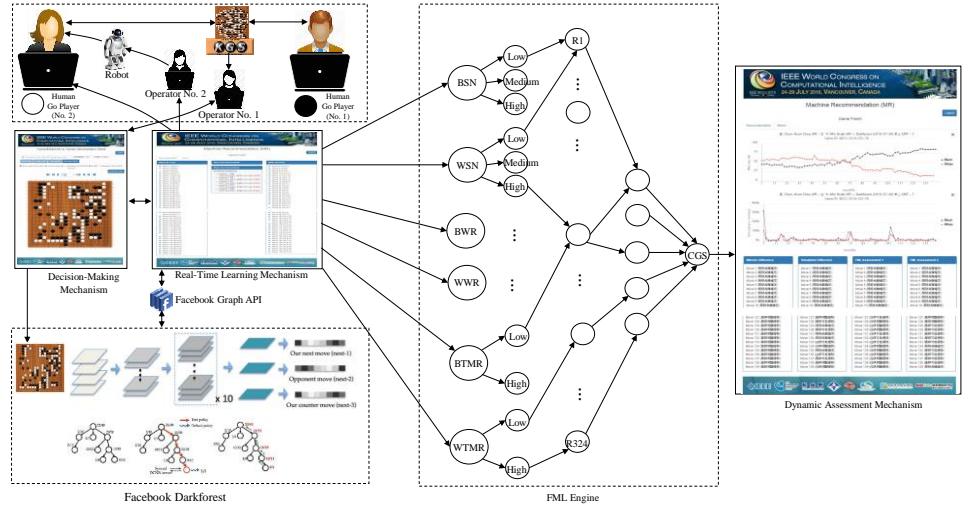

Fig. 2.     Structure of the demonstration game among humans, Darkforest, and the intelligent game bot.

Table 2.     Information on how to operate the demonstration game in the event.

| |
|---|
| **Black:** Human Go player No. 1 |
| **White:** Human Go player No. 2, computer Go program *Darkforest*, and a robot |
| **Event:** Black and White played a demonstration game via Kiseido Go Server (KGS) in the event. |
| **Step 1.** Black played move 1 on KGS. |
| **Step 2.** The operator No. 1 reflected move 1 through the *Decision-Making & Learning mechanism* to make the *Intelligent Game Bot* receive the information of move 1. |
| **Step 3.** The *Intelligent Game Bot* predicted the top-3 next moves for move 2. |
| **Step 4.** The operator No. 2 reflected the positions of the predicted top-3 next moves for move 2 to the robot through the Internet. Then, the robot spoke the received contents to White in Japanese for a reference. In addition to positions, the robot also cheered up for White or asked her to relax by sipping a cup of tea during the event. |
| **Step 5.** White referred to the predicted top-3 next moves of move 2 through the *Real-Time Learning mechanism* and also referred to the report from the robot to play her move 2 on KGS. |
| **Step 6.** The operator No. 1 reflected move 2 through the *Decision-Making & Learning mechanism* to make *Intelligent Game Bot* receive the position of move 2. |
| **Step 7.** Repeat **Steps 1 to 6** to play next move until the end of the game. |
| **Step 8.** End |
| **Note:** If the total number of moves is more than 10 moves, then the *Dynamic Assessment mechanism* starts to infer the up-to-now current game situation according to the information of number of MCTS simulations (***BSN***, ***WSN***), win rate (***BWR***, ***WWR***), and top-move rate (***BTMR***, ***WTMR***). Then, the inferred result is displayed through the *Intelligent Decision & Learning mechanism* and the *Proximal Development Agent*. Additionally, the variances in simulation number of MCTS and win rate for each move are also displayed. |

### 3.3. *Intelligent Game Bot*

In this paper, the open-source code of Facebook's Darkforest is applied to train the DCNN to predict the next top-$k$ moves to construct an FDAA. Developed by Facebook AI Research (FAIR), Darkforest is a Go game engine powered by deep learning.[4] It treats the 19×19 board as a 19×19 image with multiple channels. Each channel encodes a different aspect of board information, and features extracted from the current board situation are used as the network inputs. Figure 3 shows the adopted 12-layered ($d = 12$) full convolutional network structure of Darkforest[4] with the representational state transfer (RESTful) API mechanism and the intelligent decision-making and learning mechanism. It is described as follows: (1) the input is the current board situation with history information, and the output is a prediction of the next $k$ moves; (2) each convolution layer is followed by a rectifier linear unit nonlinearity; (3) except for the first layer, all layers use the same width ($w = 384$); and (4) Darkforest uses only one softmax layer to predict the next move by reducing the number of parameters.[4]

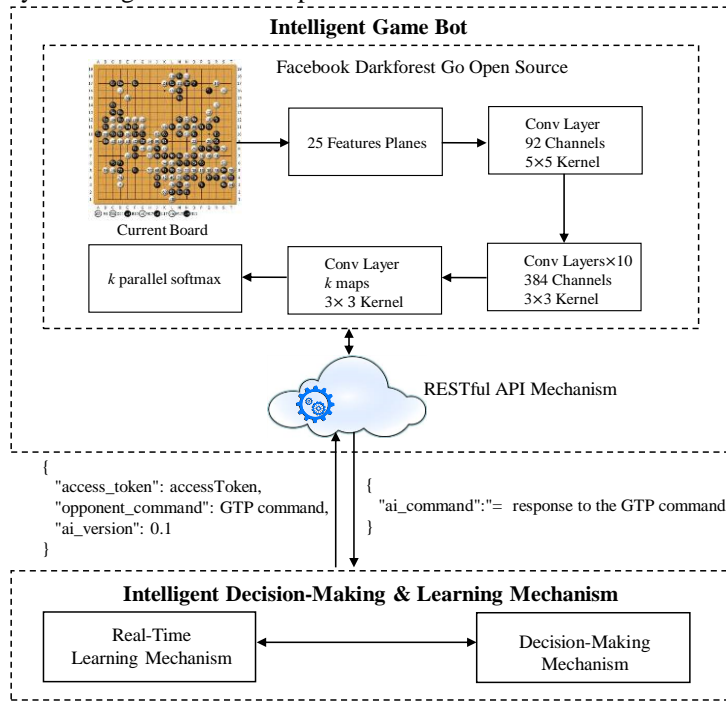

Fig. 3.    Adopted 12-layered full convolutional network structure of *Darkforest*[4] and communication between the intelligent decision-making and learning mechanism and the intelligent game bot.

According to Ref. 4, when training the policy network, Darkforest uses 16 CPU threads to prepare a minibatch. Each minibatch for each thread randomly selects one game out of 300, simulates one step according to the game record, and then extracts features and the next $k$ moves as the input/output pair in the batch. If the game has ended or fewer than $k$ moves are left, it randomly selects one replacement from the training set and continues. The batch size is 256. Darkforest randomly initializes games into different stages before training, thus avoiding quickly overfitting and becoming trapped in poor local minima.[4] It defines an epoch as 10000 minibatches and uses vanilla stochastic gradient descent on four NVidia K40m graphics processing units (GPUs) in a single machine to train the entire network. Each epoch lasts approximately 5–6 hours. The learning rate is initially 0.05, and this is divided by 5 when convergence stalls.[4]

Combining the DCNN with MCTS requires considerable effort because each rollout of MCTS is considerably faster than the DCNN evaluation. Darkforest's implementation of MCTS provides 16k rollouts per second (for 16 threads on a machine with an Intel Xeon

CPU E5-2680 v2 at 2.80GHz) while it typically takes 0.2 seconds for the DCNN to produce board evaluations of a batch size of 128 with four GPUs.[4] In synchronized implementations, MCTS waits until the DCNN evaluates the board situation of a leaf node and then expands the leaf. The default policy may run before/after the DCNN evaluation. Darkforest evaluates the synchronized case and achieves an 84.8% win rate against its raw DCNN player with only 1000 rollouts.[4]

In the system described in this paper, the Facebook Graph API adopted at the IEEE WCCI 2016 is replaced with the developed RESTful API mechanism. Figure 3 shows the communication between the intelligent decision-making and learning mechanism and the intelligent game bot. In Fig 3, the JavaScript Object Notation (JSON) is used as the data-interchange language. Take the GTP command "play white F7," for an example. When the RESTful API mechanism receives the GTP command, it passes the JSON-based data-interchange text to Darkforest. Darkforest responds "true" when the situation is appropriate for the command "play white F7." Finally, the RESTful API mechanism sends the response to the intelligent decision-making and learning mechanism.

## 4. Proximal Development Agent and Intelligent Robot Agent

In this section, the proximal development agent is described, including the adopted fuzzy variables for the dynamic assessment mechanism, the constructed FML engine, and the developed GoSocket mechanism. Then, the intelligent agent is introduced.

### 4.1. *Fuzzy Variables for Dynamic Assessment Mechanism*

In this paper, we define six input fuzzy variables and one output fuzzy variable for the *dynamic assessment mechanism*. The input fuzzy variables include the simulation number for black (**BSN**), simulation number for white (**WSN**), win rate for black (**BWR**), win rate for white (**WWR**), top-move rate for black (**BTMR**), and top-move rate for white (**WTMR**). The output fuzzy variable is defined as the current game situation (**CGS**). The proximal development agent receives the real-time number of MCTS simulations and the win rate of the predicted top-five next moves through the GoSocket mechanism. The GoSocket mechanism stores the received data in the database; meanwhile, it also selects the exact simulation number of MCTS (**SN**) and win rate (**WR**) according to the predicted top-five next moves and the actual move position.

In Fig. 4, the predicted top-five next moves for Moves 51 and 52 from Darkforest are displayed in the ascending order of *SN*. In Fig. 4, the predicted top-five next moves for Moves 51 and 52 from Darkforest are displayed in the ascending order of *SN*. The information of Move 51 in Fig. 4(a) is as follows: Suggest-1: **B1** ($BSN_1$ = 12983 and $BWR_1$ = 0.46114); Suggest-2: **H3** ($BSN_2$ = 2811 and $BWR_2$ = 0.42173); Suggest-3: **C1** ($BSN_3$ = 1813 and $BWR_3$ = 0.40786); Suggest-4 is **G2** ($BSN_4$ = 835 and $BWR_4$ = 0.41851); and Suggest-5: **F1** ($BSN_5$ = 712 and $BWR_5$ = 0.35764). From the actual situation, we know that Black played Move 51 at **B1** and White played Move 52 at **G2**. Hence, Darkforest exactly predicted the location of Moves 51 and 52. The exact values of *BSN* and *BWR* for Move 51 are 12983 and 0.46114, respectively. Likewise, the exact values of *WSN* and *WWR* for Move 52 are 13877 and 0.53501, respectively. *BTMR* and *WTMR* represent the top-five-move effect on the current game situation for Black and White, respectively. Consider the $k$th move $M_k$ within the $p$th game $G_p$. Equations (1) and (2) are used to compute $BTMR(G_p, M_k)$ and $WTMR(G_p, M_k)$, respectively.

$$BTMR\ (G_p, M_k) = \sum_{j=1}^{\lfloor\frac{k}{2}\rfloor} \left\lfloor \left( \sum_{i=1}^{5} \frac{x_i}{\lfloor\frac{k}{2}\rfloor} \times w_i - \frac{x_6}{\lfloor\frac{k}{2}\rfloor} \times w_6 \right) \times 100 \right\rfloor \qquad (1)$$

$$WTMR\ (G_p, M_k) = \sum_{j=1}^{\lfloor\frac{k}{2}\rfloor} \left\lfloor \left( \sum_{i=1}^{5} \frac{x_i}{\lfloor\frac{k}{2}\rfloor} \times w_i - \frac{x_6}{\lfloor\frac{k}{2}\rfloor} \times w_6 \right) \times 100 \right\rfloor \qquad (2)$$

where (1) $x_1$, $x_2$, $x_3$, $x_4$, and $x_5$ denote the cumulative counts of the numbers of when Darkforest correctly predicts Suggest-1, Suggest-2, Suggest-3, Suggest-4, and Suggest-5 until the $k$th move for the $p$th game, respectively; (2) $x_6$ denotes the cumulative counts of the numbers of when none of Suggest-1, Suggest-2, Suggest-3, Suggest-4, and Suggest-5 are predicted by Darkforest; an (3) $w_i$ denotes the given weight for $x_i$ where $1 \le i \le 6$. In this paper, three parameter combinations are used to define $w_i$, including: (a) *BTMR#1* or *WTMR#1* ($w_1 = 1$, $w_2 = 1$, $w_3 = 1$, $w_4 = 1$, $w_5 = 1$, and $w_6 = 0$); (b) *BTMR#2* or *WTMR#2* ($w_1 = 1$, $w_2 = 0.8$, $w_3 = 0.6$, $w_4 = 0.4$, $w_5 = 0.2$, and $w_6 = 0.1$); and (c) *BTMR#3* or *WTMR#3* ($w_1 = 1$, $w_2 = 0.8$, $w_3 = 0.6$, $w_4 = 0.4$, $w_5 = 0.2$, and $w_6 = -0.1$).

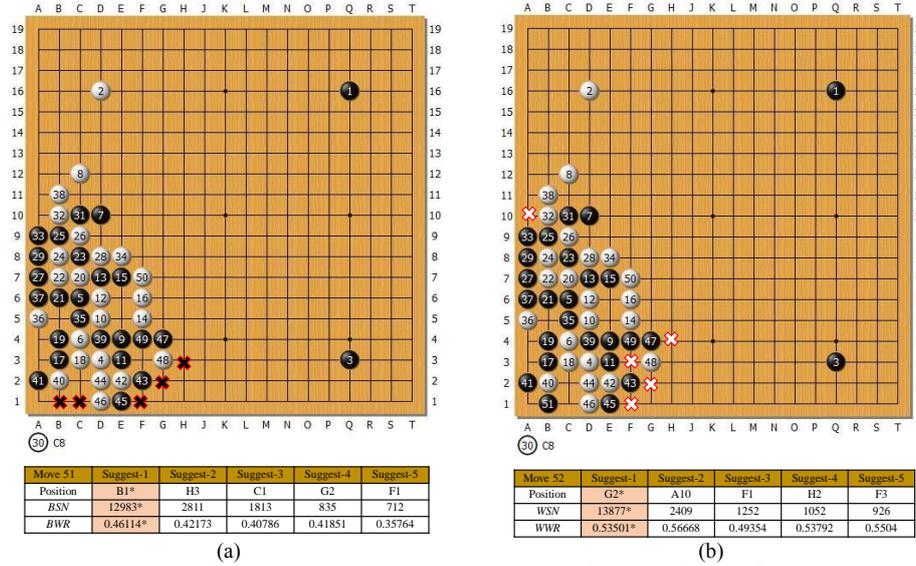

| Move 51 | Suggest-1 | Suggest-2 | Suggest-3 | Suggest-4 | Suggest-5 |
|---|---|---|---|---|---|
| Position | B1* | H3 | C1 | G2 | F1 |
| BSN | 12983* | 2811 | 1813 | 835 | 712 |
| BWR | 0.46114* | 0.42173 | 0.40786 | 0.41851 | 0.35764 |

(a)

| Move 52 | Suggest-1 | Suggest-2 | Suggest-3 | Suggest-4 | Suggest-5 |
|---|---|---|---|---|---|
| Position | G2* | A10 | F1 | H2 | F3 |
| WSN | 13877* | 2409 | 1252 | 1052 | 926 |
| WWR | 0.53501* | 0.56668 | 0.49354 | 0.53792 | 0.5504 |

(b)

Fig. 4.    Predicted top-five next moves for moves (a) 51 and (b) 52 from *Darkforest*.

Table 3 and Fig. 4(a) give a *BTMR* example for Move 51. Next, the dynamic assessment mechanism infers the performance and presents the results using the output fuzzy variable ***CGS***. The mechanism infers the game situation and obtains the linguistic results from the fuzzy sets of ***CGS***, such as "Black has an obvious advantage (***BlackObviousAdvantage***)," "Black has a possible advantage (***BlackPossibleAdvantage***)," "White has an obvious advantage (**WhiteObviousAdvantage**)," "White has a possible advantage (**WhitePossibleAdvantage**)," or "both are in an uncertain situation (**UncertainSituation**)."

Table 3.   *BTMR* example for Move 51.

| Predicted top-5 moves | Matched | | | | | Mismatched |
|---|---|---|---|---|---|---|
| | Suggest-1 | Suggest-2 | Suggest-3 | Suggest-4 | Suggest-5 | |
| $x_i$ | 11 | 5 | 6 | 1 | 0 | 3 |
| $x_i / N$ | 0.423 | 0.1923 | 0.2307 | 0.0384 | 0 | 0.1153 |
| *BTMR#1* = [(0.423 + 0.1923 + 0.2307 + 0.0384 + 0) + 0] × 100 = 88.44% | | | | | | |
| *BTMR#2* = [(0.423 + 0.15384 + 0.13842 + 0.01536 + 0) + 0.01153] × 100= 72.01% | | | | | | |
| *BTMR#3* = [(0.423 + 0.15384 + 0.13842 + 0.01536 + 0) - 0.01153] × 100= 71.9% | | | | | | |

## 4.2. *FML Engine and GoSocket Mechanism*

In this paper, we apply the basic statistics and the concept of ZPD to design the fuzzy sets and fuzzy variables. Table 4 shows the calculated values of minimum, average, maximum, and standard deviation of **SN**, **WR**, and top-move rate (**TMR**) according to the first 48 collected games, where (1) $STD_1$ denotes the standard deviation between the minimum and the mean, (2) $STD_2$ denotes the standard deviation between the mean and the maximum, and (3) $STD_3$ denotes the standard deviation for all of the data after removing the outliers. Then, we refer to Table 4 to construct the knowledge base for the *Dynamic Assessment mechanism* and the FML engine. Figure 5(a)–5(d) shows the adopted fuzzy sets for the fuzzy variables **BSN** and **WSN**, **BWR** and **WWR**, **BTMR** and **WTMR**, and **CGS**, respectively. The rule base is constructed by assigning a weight value for each input fuzzy variable and the corresponding linguistic terms listed in Table 5. Herein, we consider *BSN* and *WSN* the most crucial factor; therefore, these are given a higher weight than the other input fuzzy variables. The proposed method is detailed in a previous study.[27] Table 6 shows the adopted partial fuzzy rules, and Table 7 lists part of the knowledge base and rule base of the dynamic assessment mechanism.

Table 4.   Statistic data of fuzzy variables **SN**, **WR**, and **TMR**.

| Fuzzy Variable | Minimum | Mean | Maximum | Standard Deviation | | |
|---|---|---|---|---|---|---|
| | | | | $STD_1$ | $STD_2$ | $STD_3$ |
| **BSN / WSN** | 3420 | 9883 | 14999 | 2762 | 1421.56 | 1450.62 |
| **BWR / WWR** | 0.2 | 0.49 | 0.6 | 0.09 | 0.07 | 0.05 |
| **BTMR / WTMR** | 0 | 0.382 | 0.5 | 0.11 | 0.09 | 0.03 |

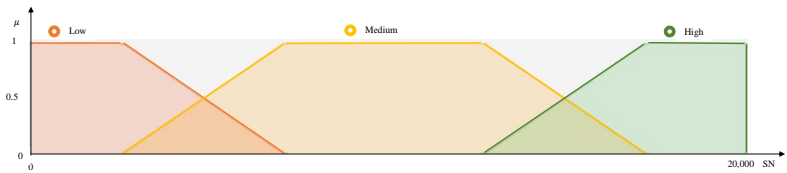

(a)

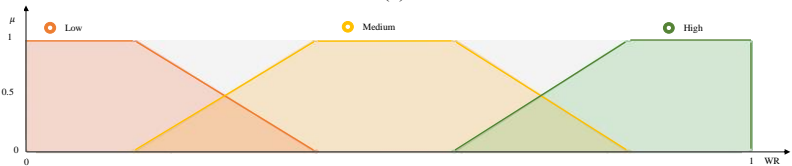

(b)

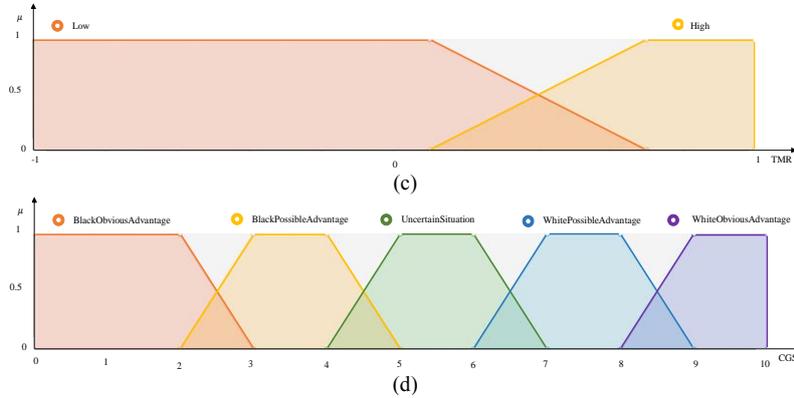

Fig. 5.    Fuzzy sets (a) ***BSN*** / ***WSN***, (b) ***BWR*** / ***WWR***, (c) ***BTMR*** / ***WTMR***, and (d) ***CGS***.

Table 5.    The fuzzy relation weight of each input fuzzy variable and its related fuzzy term weight.

| No | Fuzzy Variable | *Fuzzy Relation Weight* | Fuzzy Term (Weight) | | |
|---|---|---|---|---|---|
| 1 | ***BSN*** | 2.5 | *Low* (3) | *Medium* (2) | *High* (1) |
| 2 | ***WSN*** | | *Low* (1) | *Medium* (2) | *High* (3) |
| 3 | ***BWR*** | 2 | *Low* (3) | *Medium* (2) | *High* (1) |
| 4 | ***WWR*** | | *Low* (1) | *Medium* (2) | *High* (3) |
| 5 | ***BTMR*** | 1 | *Low* (2) | *High* (1) | N/A |
| 6 | ***WTMR*** | | *Low* (1) | *High* (2) | N/A |

Table 6.    Partial fuzzy rules.

| No. | Input Fuzzy Variables | | | | | | Output Fuzzy Variable |
|---|---|---|---|---|---|---|---|
| | ***BSN*** | ***WSN*** | ***BWR*** | ***WWR*** | ***BTMR*** | ***WTMR*** | ***CGS*** |
| 1 | Low | Low | Low | Low | Low | Low | **UncertainSituation** |
| 2 | Low | Low | Low | Low | Low | High | **UncertainSituation** |
| 3 | Low | Low | Low | Low | High | Low | **UncertainSituation** |
| 4 | Low | Low | Low | Low | High | High | **UncertainSituation** |
| 5 | Low | Low | Low | Medium | Low | Low | **UncertainSituation** |
| 6 | Low | Low | Low | Medium | Low | High | **WhitePossibleAdvantage** |
| 7 | Low | Low | Low | Medium | High | Low | **UncertainSituation** |
| 8 | Low | Low | Low | Medium | High | High | **UncertainSituation** |
| 9 | Low | Low | Low | High | Low | Low | **WhitePossibleAdvantage** |
| 10 | Low | Low | Low | High | Low | High | **WhitePossibleAdvantage** |
| | | | | ⋮ | | | |
| 315 | High | High | High | Low | High | Low | **BlackPossibleAdvantage** |
| 316 | High | High | High | Low | High | High | **BlackPossibleAdvantage** |
| 317 | High | High | High | Medium | Low | Low | **BlackPossibleAdvantage** |
| 318 | High | High | High | Medium | Low | High | **UncertainSituation** |
| 319 | High | High | High | Medium | High | Low | **BlackPossibleAdvantage** |
| 320 | High | High | High | Medium | High | High | **BlackPossibleAdvantage** |
| 321 | High | High | High | High | Low | Low | **UncertainSituation** |
| 322 | High | High | High | High | Low | High | **UncertainSituation** |
| 323 | High | High | High | High | High | Low | **UncertainSituation** |
| 324 | High | High | High | High | High | High | **UncertainSituation** |

Table 7.    Partial knowledge base and rule base of the dynamic assessment mechanism.

```
<?xml version="1.0" encoding="UTF-8"?>
<fuzzySystem xmlns="http://www.ieee1855.org" name="GameSystem" networkAddress="127.0.0.1">
    <knowledgeBase networkAddress="127.0.0.1">
        <fuzzyVariable name="BSN" scale="" domainleft="0" domainright="20000" type="Input"
accumulation="MAX" defuzzifier="COG" defaultValue="0.0" networkAddress="127.0.0.1">
```

```xml
            <fuzzyTerm name="Low" complement="false">
                <trapezoidShape param1="0" param2="0" param3="2556" param4="7122"/>
            </fuzzyTerm>
            <fuzzyTerm name="Medium" complement="false">
                <trapezoidShape param1="2556" param2="7122" param3="12637" param4="17203"/>
            </fuzzyTerm>
            <fuzzyTerm name="High" complement="false">
                <trapezoidShape param1="12637" param2="17203" param3="20000" param4="20000"/>
            </fuzzyTerm>
        </fuzzyVariable>
...
    </knowledgeBase>
<mamdaniRuleBase name="ruleBase1" activationMethod="MIN" andMethod="MIN" orMethod="MAX"
networkAddress="127.0.0.1">
<rule name="rule-1" andMethod="MIN"                  <rule name="rule-324" andMethod="MIN"
orMethod="MAX" connector="AND" weight="1.0"         orMethod="MAX" connector="AND" weight="1.0"
networkAddress="127.0.0.1">                          networkAddress="127.0.0.1">
        <antecedent>                                         <antecedent>
            <clause>                                             <clause>
                <variable>BSN</variable>                             <variable>BSN</variable>
                <term>Low</term>                                     <term>High</term>
            </clause>                                            </clause>
            <clause>                                             <clause>
                <variable>WSN</variable>                             <variable>WSN</variable>
                <term>Low</term>                                     <term>High</term>
            </clause>                                            </clause>
            <clause>                                             <clause>
                <variable>BWR</variable>                             <variable>BWR</variable>
                <term>Low</term>                                     <term>High</term>
            </clause>                                            </clause>
            <clause>                                             <clause>
                <variable>WWR</variable>                             <variable>WWR</variable>
                <term>Low</term>                                     <term>High</term>
            </clause>                                            </clause>
            <clause>                                             <clause>
                <variable>BTMR</variable>                            <variable>BTMR</variable>
                <term>Low</term>                                     <term>High</term>
            </clause>                                            </clause>
            <clause>                                             <clause>
                <variable>WTMR</variable>                            <variable>WTMR</variable>
                <term>Low</term>                                     <term>High</term>
            </clause>                                            </clause>
        </antecedent>                                         </antecedent>
        <consequent>                                          <consequent>
            <then>                                               <then>
                <clause>                                             <clause>
                    <variable>CGS</variable>                             <variable>CGS</variable>
                    <term>UncertainSituation</term>                     <term>UncertainSituation</term>
                </clause>                                            </clause>
            </then>                                              </then>
        </consequent>                                         </consequent>
    </rule>                                               </rule>
...
</mamdaniRuleBase>
</fuzzySystem>
```

### 4.3. *Intelligent Robot Agent*

The intelligent agent[28] includes a GoSocket engine, a knowledge base, and a summarization agent. The GoSocket engine communicates with the proximal development agent and then stores the received data in the knowledge base. The intelligent agent can indicate the predicted top-three next moves, summarize the game with linguistic descriptions to generate a short commentary, and finally send the related information to the robot through the Internet to report to the game players through their computer or social media platform. Table 8 shows a template of the short commentary for the summarized results of one game, and overall game situation (*OGS*) may be "favorable to Black," "favorable to White," or "uncertain situation."

Table 8.    Template of the overall game comments.

| |
|---|
| Summarization on this game given by the *Intelligent Robot Agent* is as follows:<br>**Black:** The first 3 highest simulation numbers occurred at Moves **BHighM1**(*BSN#1*), **BHighM2**(*BSN#2*), and **BHighM3**(*BSN#3*). The last 3 lowest simulation numbers occurred at Moves **BLowM1**(*BSN#4*), **BLowM2**(*BSN#5*), and **BLowM3**(*BSN#6*). The information of estimated possible win rate: The highest win rate is **BHighWR**(*BWR#1*), the lowest win rate is **BLowWR**(*BWR#2*), and the average win rate is *BAveWR#3*. Top-move rate is *BTMR#1*.<br>**White:** The first 3 highest simulation numbers occurred at Moves **WHighM1**(*WSN#1*), **WHighM2** (*WSN#2*), and **WHighM3**(*WSN#3*). The last 3 lowest simulation numbers occurred at Moves **WLowM1** (*WSN#4*), **WLowM2**(*WSN#5*), and **WLowM3**(*WSN#6*). The information of estimated possible win rate: The highest win rate is **WHighWR**(*WWR#1*), the lowest win rate is **WLowWR**(*WWR#2*), and the average win rate is *WAveWR#3*. Top-move rate is *WTMR#1*.<br>**Overall** game situation is *OGS*.<br><br>Note:<br>• (**BHightM1**, **BHightM2**, and **BHightM3**) and (**WHightM1**, **WHightM2**, and **WHightM3**) denote the Black and White move number with the first, second, and third highest simulation numbers, respectively.<br>• (*BSN#1*, *BSN#2*, and *BSN#3*) and (*WSN#1*, *WSN#2*, and *WSN#3*) denote the first, second, and third highest values of simulation numbers for Black and White, respectively.<br>• (**BLowM1**, **BLowM2**, and **BLowM3**) and (**WLowM1**, **WLowM2**, and **WLowM3**) denote the Black and White move numbers with the first, second, third lowest simulation numbers, respectively.<br>• (*BSN#4*, *BSN#5*, and *BSN#6*) and (*WSN#4*, *WSN#5*, and *WSN#6*) denote the first, second, and third lowest values of simulation numbers for Black and White, respectively.<br>• **BHighWR** and **WHighWR** denote the Black and White move numbers with the highest win rate, respectively.<br>• **BLowWR** and **WLowWR** denote the Black and White move numbers with the lowest win rate, respectively.<br>• (*BWR#1* and *BWR#2*) and (*WWR#1* and *WWR#2*) denote the highest and lowest win rates of Black and White, respectively.<br>• *BAveWR#3* and *WAveWR#3* denote the average win rate of Black and White, respectively.<br>• *BTMR#1* and *WTMR#1* denote the top-move rates of Black and White, respectively, by setting $w_1$ to $w_5$ to 1 and $w_6$ to 0 listed in Equations (1) and (2).<br>• *OGS* denotes the overall game result in linguistics which could be "*Favorable to Black*," "*Favorable to White*," or "*Uncertain Situation*." |

## 5.  Experimental Results

The developed FHMCS was mainly implemented with PHP and Java by the *Center for Research of Knowledge Application and Web Service* (KWS Center) at the *National University of Tainan* (NUTN), Taiwan. The robot PALRO[28] was used in cooperation with the Kubota Laboratory of *Tokyo Metropolitan University* (TMU), Japan. In addition, the open-source code for Darkforest was provided by the computer Go team of FAIR, USA.

We invited human Go players to play more than 150 games from May to November in 2016. On the basis of the overall game situation from the proposed approach and the known game results, some experimental results are shown in this section. We first used 60 games (Google Master vs. professional Go players in December 2016 and January 2017)[40] to verify and validate the performance of various types of the proposed FHMCS at the *National Center of High Performance Computing* (NCHC) and NUTN in Taiwan. Next, five selected games, including game records on the Pair Go World Cup 2016 Tokyo and human vs. computer competitions at the IEEE WCCI 2016 and ICIRA 2016, were adopted to observe the predicted accuracy of FHMCS on the FAIR server (**FHMCS-FAIR**) in the United States.

### 5.1. *Collected Games and Experiment Descriptions*

Table 9 shows the information of the selected 65 games in this paper. We designed various application scenarios, including **Part I** (1)**:** human vs. Google's Master Go for Games 1–60 played from December 29, 2016, to January 4, 2017,[40] and (2) **Part II:** (1) human vs. Darkforest for Game 61; (2) human vs. human for Games 62 and 63; and (3) human vs. (human + Darkforest + intelligent game bot) for Games 64 and 65. Table 10 gives the descriptions of the experiments from Part I. Servers with four GPUs on FHMCS-1 and FHMCS-2 were provided by NCHC. In addition, a server with two GPUs on FHMCS-3 was provided by NUTN.

Table 9.    Information of Games 1 to 65.

| Game | Date / Event | Black | White | Winner |
|------|-------------|-------|-------|--------|
| **Part I: FML-based Human-Machine Cooperative System (FHMCS)** | | | | |
| 1-60 | Dec. 29, 2016 to Jan. 4, 2017 | *Master* vs. Professional Go Players on the online servers Tygem and FoxGo | | *Master* |
| **Part II: FML-based HMCS on FAIR server (FHMCS-FAIR) in USA** | | | | |
| *Human vs. Darkforest Computer Go Program* | | | | |
| 61 | May 31, 2016 NUTN Testing | FB *Darkforest* | Shang-Rong Tsai (6D) | Black |
| *Human vs. Human* | | | | |
| 62 | Jul. 9, 2016 Pair Go Tokyo 2016 | Jeong Choi (6P)+ Jeonghwan Park (9P) | Yi-Min Hsieh (6P) + Iyama Yuta (9P) | Black |
| 63 | Jul. 14, 2016 Pair Go Tokyo 2016 | Shih-Luan Chen (9P) + Joanne Missingham (7P) | Jie Ke (9P) + Zhiying Yu (5P) | White |
| *Human vs. (Human + Computer Go Program Darkforest + Robot)* | | | | |
| 64 | Jul. 25, 2016 IEEE WCCI 2016 | Yi-Min Hsieh (6P) + FB *Darkforest* + Robot | Chun-Hsun Chou (9P) | White |
| 65 | Aug. 23, 2016 ICIRA 2016 | Yi-Min Hsieh (6P) + FB *Darkforest* + Robot | Chun-Hsun Chou (9P) | Uncertain |

The application scenarios from Experiments 1 to 4 are described as follows: First, login to the developed FHMCS for Go, as shown in Fig. 6. Second, click on the "add game" button to add a game by setting the number of MCTS simulations, komi, and play mode. Third, click on the "load multi-SGF files" button to selected games and upload their SGF files. Fourth, the developed FHMCS simulates the game play according to the uploaded SGF files and their settings. Fifth, after the game has been completed, the predicted information is obtained from Darkforest and the results of the win rate and top-move rate are compared among Experiments 1–4. Finally, the FDAA infers the game results to validate the accuracy of the proposed approach.

Table 10. Experiment descriptions of Part I.

| Experiment | Number of Simulations Setting / FHMCS | GPU Number / Location | GPU Card Model |
|---|---|---|---|
| 1 | 20000 / FHMCS-1 | **4 GPUs** / NCHC | Tesla K80 × 2 |
| 2 | 20000 / FHMCS-2 | **4 GPUs** / NCHC | GForce GTX 1080× 4 |
| 3 | 10000 / FHMCS-1 | **4 GPUs** / NCHC | Tesla K80 × 2 |
| 4 | 10000 / FHMCS-2 | **4 GPUs** / NCHC | GForce GTX 1080× 4 |
| 5 | 3000 / FHMCS-1 | **4 GPUs** / NCHC | Tesla K80 × 2 |
| 6 | 3000 / FHMCS-2 | **4 GPUs** / NCHC | GForce GTX 1080× 4 |
| 7 | 3000 / FHMCS-3 | **2 GPUs** / NUTN | Quadro K2200 × 1 Quadro M2000 × 2 |
| 8 | 1500 / FHMCS-1 | **4 GPUs** / NCHC | Tesla K80 × 2 |
| 9 | 1500 / FHMCS-2 | **4 GPUs** / NCHC | GForce GTX 1080× 4 |
| 10 | 1500 / FHMCS-3 | **2 GPUs** / NUTN | Quadro K2200 × 1 Quadro M2000 × 2 |
| Note | FHMCS-1, FHMCS-2, and FHMCS-3 denote that the developed FML-based Human-Machine Cooperative System are operated under the server whose GPU number / Location (GPU card model) is 4GPUs / NCHU (Tesla K80 × 2), 4GPUs / NCHU (GForce GTX 1080× 4), and 2 GPUs / NUTN (Quadro K2200 × 1 and Quadro M2000 × 2), respectively. | | |

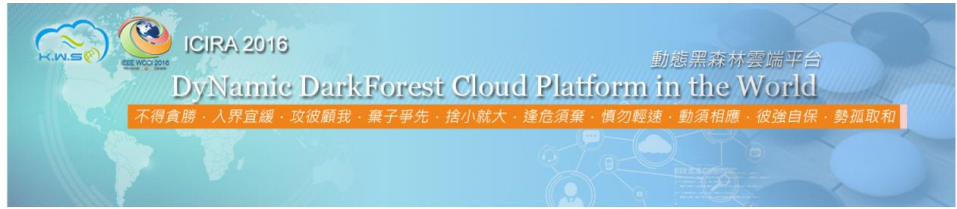

Human&Machine vs. Human Demonstration Game

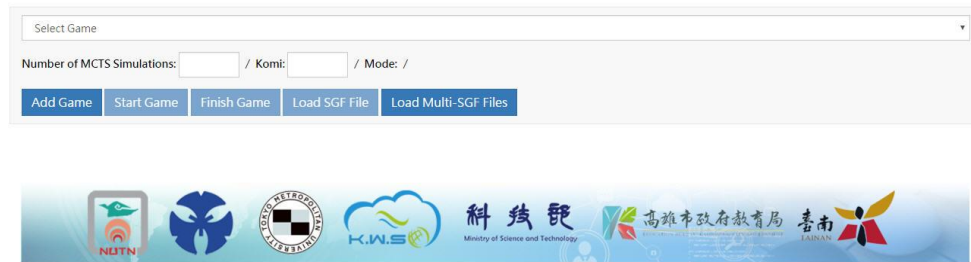

Fig. 6.     Screenshot of the developed FHMCS for game of Go.

## 5.2. Comparison Results of WR and TMR values from Experiments 1 to 4

To evaluate the performance under different server specifications, we compared *WR* for machines and humans, respectively. We first logged in to the developed system and uploaded the SGF file for Game 1 onto the constructed system with GPU cards (two Tesla K80 GPUs or four GForce GTX 1080 GPUs) to simulate the game play. Next, we set the number of simulations to 20000 or 10000 to run the game play through to its completion. Third, we repeated these steps for the remaining games. Finally, we acquired the machines' and humans' last move of FDAA-predicted *WR* for Games 1–60. Figure 7(a) shows the machines' and humans' FDAA-predicted *WR* values derived from Experiments 1 and 2. Figure 7(b) shows the corresponding results from Experiments 3 and 4. Figure 7 shows that the FDAA-predicted *WR* is higher for the machines than for the humans in most games. The exceptions include Games 29 (the machine won by 0.5 points, Experiments 1 and 3), Game 46 (the machine won by resignation, Experiment 2), and Game 22 (the machine won by 4.5 points, Experiments 2 and 4). These results show that the FDAA can correctly predict most of the games. Next, we compare the *TMR* values obtained in Experiments 1–4. Figure 8(a) and 8(b) shows the machines' and humans' *TMR* values from Experiments 1 and 2, as well as Experiments 3 and 4, respectively. The figure shows that for both the machines and humans, Experiments 1 and 3 yielded markedly higher *TMR* values than did Experiments 2 and 4; therefore, the *TMR* value might be more closely related to the GPU model than the number of simulations.

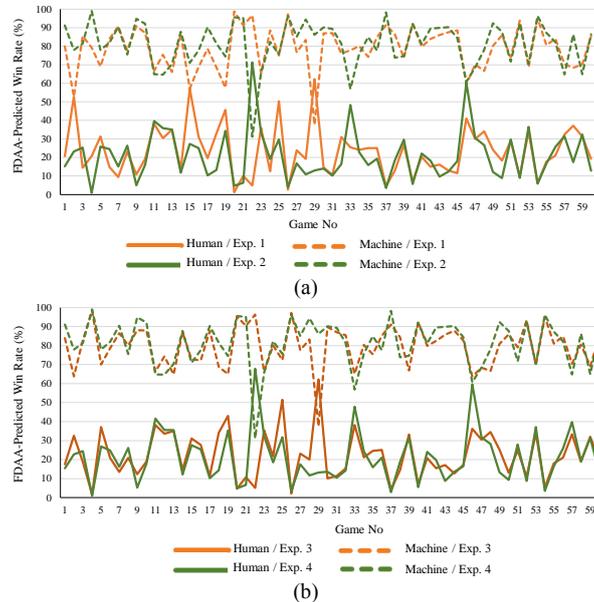

Fig. 7.    FDAA-predicted Machine and human's *WR* values derived from (1) Experiments 1–2 and (2) Experiments 3–4.

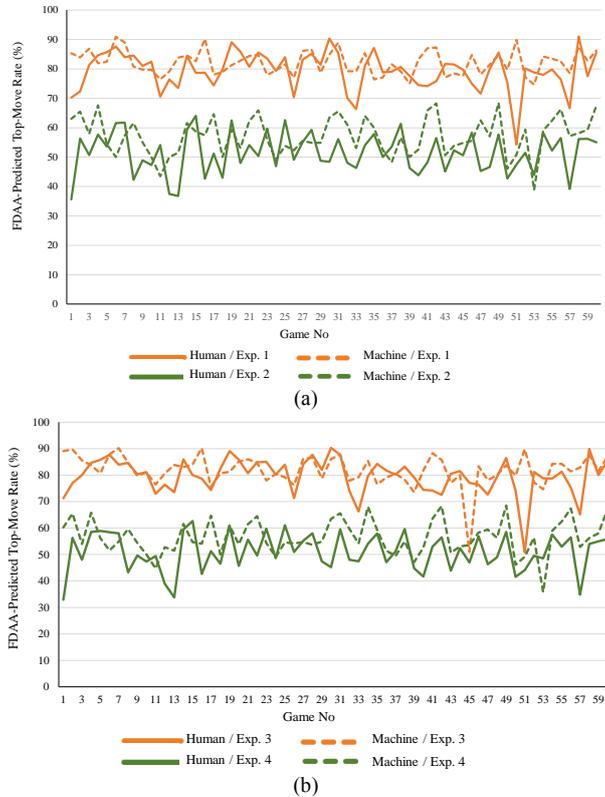

Fig. 8.    FDAA-predicted Machine and human's *TMR* values derived from (1) Experiments 1–2 and (2) Experiments 3–4.

### 5.3. *Accuracy of Dynamic Assessment Mechanism*

This subsection shows the performance of the proposed dynamic assessment mechanism, which is based on two versions of code written in FML (denoted as FML-1 and FML-2) with different decision-making methods, respectively. We use FML-1 and FML-2 to infer the *CGS* of each move, respectively. The knowledge base of FML-1 has four input fuzzy variables (FML-1 with four IFVs), namely ***BSN***, ***WSN***, ***BWR***, and ***WWR***. The second one (FML-2) has six input fuzzy variables (FML-2 with six IFVs), namely ***BSN***, ***WSN***, ***BWR***, ***WWR***, ***BTMR***, and ***WTMR***. Table 7 shows additional details on FML-2. The accuracy is decided using two decision-making methods, denoted as Method 1 and Method 2, which are briefly described in Table 11. Figure 9 shows the accuracy of the proposed approach, indicating that FML-2/Method-2 demonstrated the optimal performance and that Method 2 is more efficient than Method 1. The difference between this work and a previous study[41] is that the overall game situation of Ref. 41 is acquired on the basis of a partial game situation (e.g., three or four subgames) as well as the results of the FML assessment engine and FML-based decision support engine.[41] In the present study, the overall game situation is decided by the proposed methods listed in Table 11.

Figure 10(a) and 10(b) shows the average FDAA-predicted ***TMR***, average FDAA-predicted ***WR***, and the accuracy of the proposed approach for predicting the game outcome for humans and machines from Experiments 1 to 7, respectively. The figure indicates that the proposed approach (FML-2/Method-2) attains higher accuracy than the average FDAA-predicted ***WR***. Furthermore, the average FDAA-predicted ***WR*** for machines is higher than that for humans. Additionally, Figure 9 shows that the ***TMR*** value derived from Experiments 1, 3, and 5 (two Tesla K80 GPUs) remains near 80%, regardless of whether *SN* is set to 20000, 10000, or 3000. However, in Experiments 2, 4, and 6 (four GForce GTX 1080 GPUs), the ***TMR*** remained approximately 50%. For Experiment 7, Figure 10 shows that the average FDAA-predicted ***TMR*** is also near 50%, despite only two GPUs being used for this experiment. Therefore, the average FDAA-predicted ***TMR*** is more closely related to the model of the GPU card than the number of simulations.

Table 11. Descriptions of two decision-making methods.

- **Method 1:** *Overall Game Situation* (*OGS*) is decided by the inferred *CGS* of the last move. If the *CGS* matches with the actual game result, then the proposed *Dynamic Assessment mechanism* correctly infers the *CGS*.
- **Method 2:** 1) *OGS* is decided by the inferred current game situation of the last 5 moves. If all *CGS* results of the last 5 moves are not "*Uncertain*", then judge if any move of *CGS* matches with the actual game result. If anyone matches, then the proposed *Dynamic Assessment mechanism* correctly infers the *OGS*. 2) If all *CGS* results of the last 5 moves are "*Uncertain*", then look *N* moves up until finding five "non-uncertain *CGS*." Next, we judge if any move of *CGS* matches with the actual game result. If anyone matches, then the proposed *Dynamic Assessment mechanism* correctly infers the *OGS*.

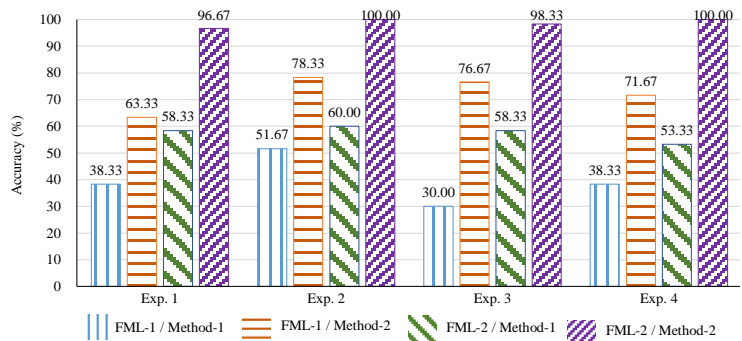

Fig. 9. Accuracy of different FMLs based on different decision-making methods.

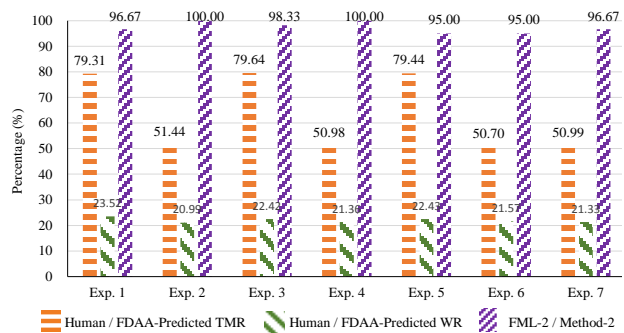

(a)

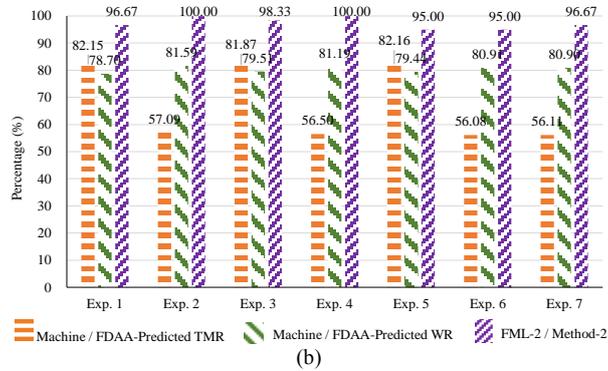

(b)

Fig. 10.    The average FDAA-predicted ***TMR***, average FDAA-predicted ***WR***, and the accuracy of the proposed approach for predicting the game outcome for (a) humans and (b) machines from Experiments 1 to 7.

### 5.4. *Behavioral Analysis in TMR/WR/accuracy based on SN*

This subsection shows the behavioral analysis based on various combinations of FHCMS and the number of simulations. Figure 11(a) shows that the semantics of the average FDAA-predicted ***TMR*** for the proposed FHCMS-1, FHCMS-2, and FHCMS-3 will be "high," "medium," and "medium," respectively, if we assume that the fuzzy set of "Percentage (%)" shown in Fig. 11(b) is the one used. Figure 11(c) shows that the human's semantics of average FDAA-predicted ***WR*** for the proposed FHCMS-1, FHCMS-2, and FHCMS-3 are all "low." However, the semantics for the machine ***WR*** are all "high" for the proposed FHCMS-1, FHCMS-2, and FHCMS-3. Additionally, Fig. 11(c) indicates that the average FDAA-predicted machine ***WR*** of FHMCS-3 with 3000 simulations is higher than the one with 1500 simulations. Figure 11(d) shows that the accuracy is "very high" for both FHCMS-1 and FHCMS-2 but "medium" for FHCMS-3 with 1500 simulations and "very high" for the FHCMS-3 with 3000 simulations.

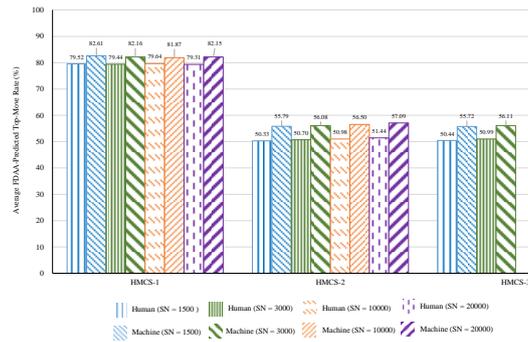

(a)

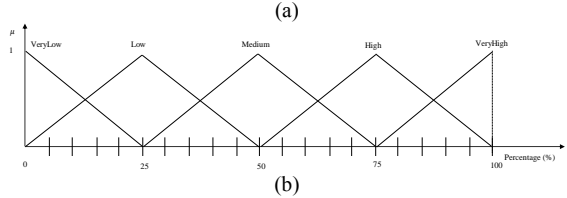

(b)

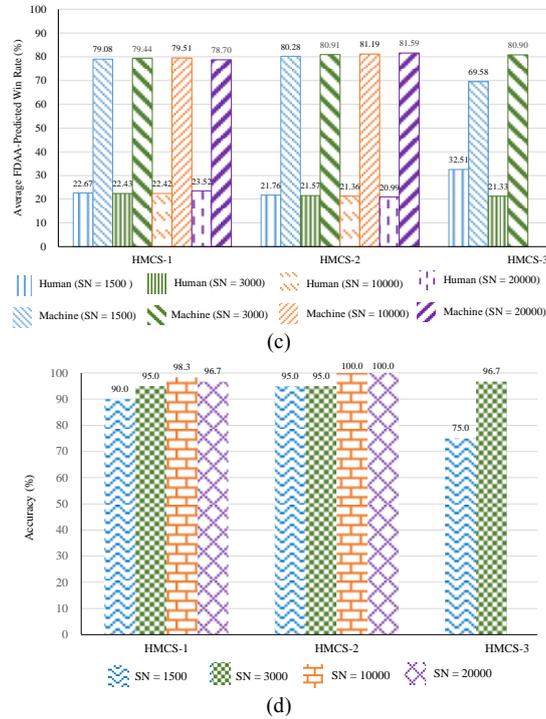

(c)

(d)

Fig. 11.    Bar charts of (a) the average FDAA-predicted **TMR**, (b) fuzzy set of *Percentage* (%), (c) the average FDAA-predicted **WR**, and (d) the proposed approach accuracy based on varous combinations of FHMCSs and **SN**.

### 5.5. *Commentary given by the Summarization Agent*

This subsection presents information on Game 51 (Google Master as White vs. Chun-Hsun Chou as Black; game result = W + R). Figure 12(a) and 12(b) respectively shows the *SN* and *WR* curves from Experiments 1 and 7. Figure 12(a) shows that *SN* occasionally exhibits sharp variance in the neighborhood. Figure 12(b) shows that White wins the game. Meanwhile, Table 12 shows the commentary on Game 51 given by the proposed summarization agent in Experiments 1 and 7.

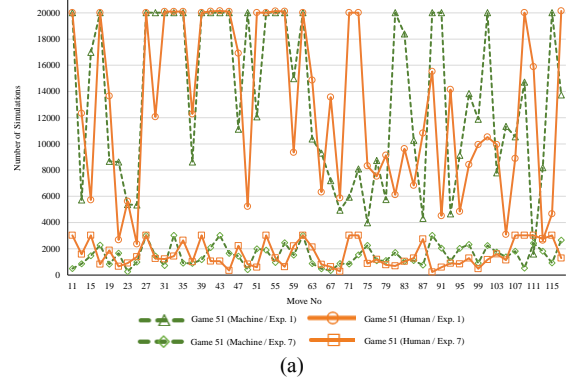

(a)

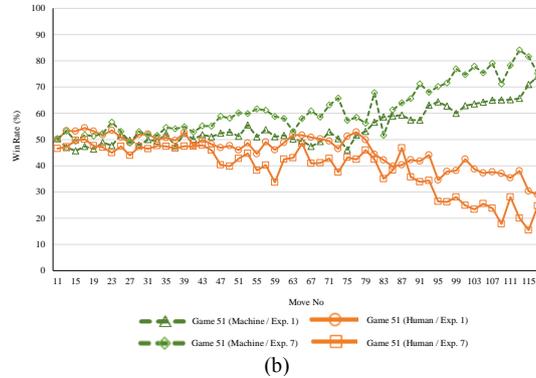

(b)

Fig. 12.     (a) Number of simulations and (b) win rate curves of Game 51.

Table 12.   Game 51: Commentary given by the summarization agent in the Experiments (a) 1 and (b) 7.

| (a) |
|---|
| **Black:** |
| – The first 3 highest simulation numbers occurred at Moves B117 (20152), B43 (20148), and B55 (20132). The last 3 lowest simulation numbers occurred at Moves B25 (2359), B113 (2635), and B21 (2678). |
| – The information of estimated possible win rate: The highest win rate is B17 (54.45%), the lowest win rate is B117 (29.01%), and the average win rate is 46.0%. |
| – Top-move rate is 54.24%. |
| **White:** |
| – The first 3 highest simulation numbers occurred at Moves W116 (20017), W102 (20017), and W92 (20017). The last 3 lowest simulation numbers occurred at Moves W112 (1606), W76 (3979), and W88 (4316). |
| – The information of estimated possible win rate: The highest win rate is W118 (74.06%), the lowest win rate is W16 (45.69%), and the average win rate is 54.65%. |
| – Top-move rate is 89.83%. |
| **Overall game situation is favorable to White.** |
| (b) |
| **Black:** |
| – The first 3 highest simulation numbers occurred at Moves B111 (3017), B109 (3017), and B107 (3017). The last 3 lowest simulation numbers occurred at Moves B89 (231), B69 (278), and B45 (342). |
| – The information of estimated possible win rate: The highest win rate is B17 (50.06%), the lowest win rate is B115 (15.49%), and the average win rate is 38.99%. |
| – Top-move rate is 47.46%. |
| **White:** |
| – The first 3 highest simulation numbers occurred at Moves W90 (3017), W62 (3017), and W44 (3017). The last 3 lowest simulation numbers occurred at Moves W24 (261), W68 (345), and W50 (401). |
| – The information of estimated possible win rate: The highest win rate is W114 (84.17%), the lowest win rate is W28 (48.68%), and the average win rate is 61.54%. |
| – Top-move rate is 50.85%. |
| **Overall game situation is favorable to White.** |

### 5.6. *Top-Move Rate (TMR) Prediction*

In this subsection, we examined the ***TMR*** prediction from Darkforest and take two games from the Pair Go World cup 2016 Tokyo[24] as an example. The game details are shown in Table 9. ***BTMR*** and ***WTMR*** are respectively 96.05% and 81.33% for Game 62 and 90.72% and 87.6% for Game 63, which shows a relatively high level of accuracy when predicting a professional Go player's next move. Additionally, we compare the ***TMR*** between professional and amateur Go players. Figure 13(a) shows a scatter diagram of ***BTMR*** and **WTMR** for the investigated games. The averages for professional and amateur Go players are 89.09% and 82.02%, respectively. Figure 13(b) shows a bar chart of the average **TMR** for 6D, 7D, 6P, 9P, and all invited Go players. The figure shows that the average **TMR** of all professional Go players is higher than that of all amateur Go players. Next, we use Game 64 to observe the **TMR**. Figure 14(a) shows that when Facebook Darkforest and the intelligent game bot are played under Suggest-1, the **BTMR** is more than 50% and the mismatch rate is less than 10%. Figure 14(b) shows that all of the White-playing moves are within the prediction of Facebook's Darkforest.

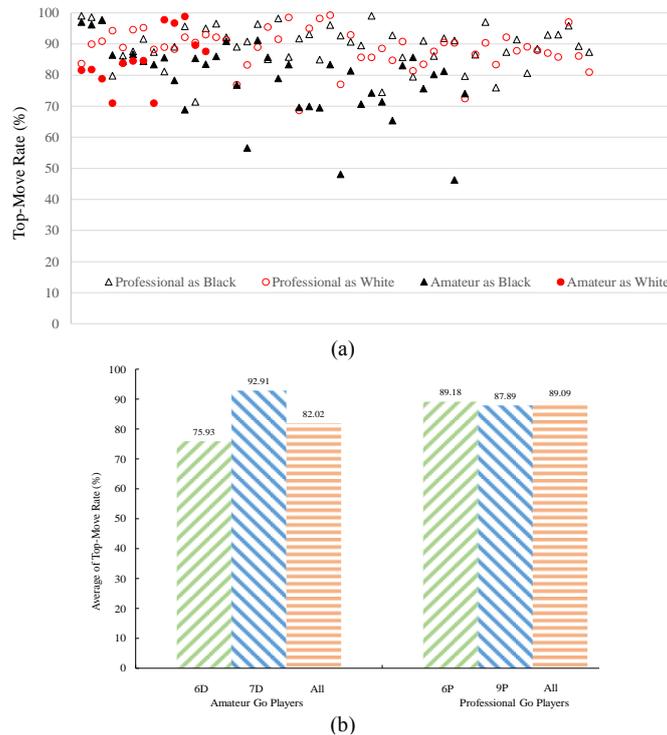

(a)

(b)

Fig. 13.    (a) Scatter diagram and (b) bar chart of top-move rate for professional and amateur Go players.

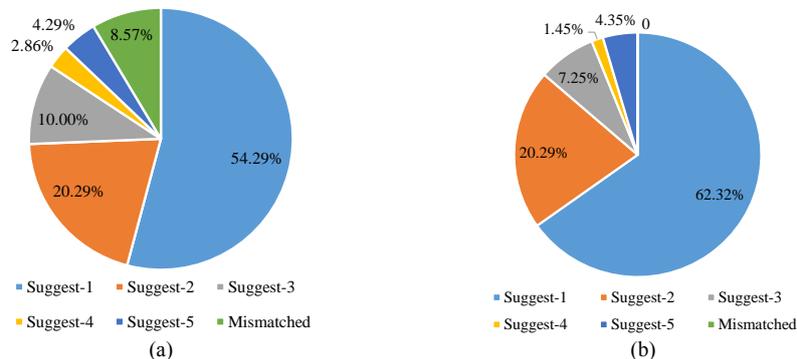

(a)                                    (b)

Fig. 14.    Pie chart of human Go player's *TMR* situation (a) *BTMR* and (b) *WTMR* of Game 64.

## 6. Conclusions

This paper presents an FHMCS that was developed by applying FML to construct a dynamic agent for the game of Go, the purpose of which is to provide Go players with information on the situation of a match at each movement. Additionally, the FML for player performance evaluation and game result commentary based on linguistic descriptions is presented. The experimental results are summarized as follows:

- The results of the experiments in **Part I (Experiments 1–10)** are as follows: (1) The proposed FDAA can correctly predict most games. (2) Regardless of whether it is for predicting the moves of machines or humans, Experiments 1 and 3 attained markedly higher *TMR* vales than did Experiments 2 and 4. (3) *TMR* may be more closely related to the GPU model than the number of simulations. (4) The accuracy values of the proposed approach (FML-2/Method-2) in Experiments 1, 2, 3, and 4 are 96.67%, 100%, 100%, and 100%, respectively. FML-2/Method-2 attained the best performance among all methods examined in this study.
- The results of the experiments in **Part II** are as follows: (1) From the invited professional Go players' viewpoints, it is helpful for Go learners to learn the game of Go together with a robot, especially for children. (2) The accuracy for predicting a professional Go player's next move is relatively high. (3) The average *TMR* of all professional Go players is higher than that of all amateur Go players.

In the future, we will include machine learning to optimize the parameters of the FML knowledge base and assign each move a different weight according to the stage of the game in order to improve the inferred result of the overall game situation. Additionally, we will develop an intelligent robot with co-learning ability for the game of Go and include languages such as English, Taiwanese, Hakka, and aboriginal languages.

**Acknowledgements**

The authors would like to thank the Ministry of Science and Technology of Taiwan for its partial support under MOST 104-2622-E-024-005-CC2, MOST 105-2622-E-024-003-CC2, and MOST 105-2221-E-024-017. We also would like to thank all invited Go players from Haifong Weiqi Academy, Taiwan for their kind help and also thank Dr. Yuandong Tian and Yan Zhu from Facebook AI Research (FAIR). Moreover, the authors would like to thank the National Center of High Performance Computing (NCHC) and Kaohsiung City Government, Taiwan for their computing resource support. Finally, the authors would like to thank IEEE CIS and general chairs of IEEE WCCI 2016 (Prof. Kan Chen Tan and Prof. Gary Yen) for their great support for human vs. computer Go competition at IEEE WCCI 2016 in Canada.